%% file: coling2020.tex
\documentclass[11pt]{article}
\usepackage{coling2020}
\usepackage{times}
\usepackage{url}
\usepackage{latexsym}
\usepackage{graphicx}
\usepackage{amsmath}
\usepackage{amssymb}
\usepackage[ruled,vlined]{algorithm2e}
\usepackage{enumitem}
\usepackage{float}
\usepackage{setspace}
\usepackage{subcaption}
\usepackage{amsfonts}
\usepackage{tabulary,booktabs}
\usepackage[acronym,nomain,nonumberlist]{glossaries}
\usepackage{multirow}
\newcolumntype{L}[1]{>{\raggedright\let\newline\\\arraybackslash\hspace{0pt}}m{#1}}
\newcolumntype{C}[1]{>{\centering\let\newline\\\arraybackslash\hspace{0pt}}m{#1}}
\newcolumntype{R}[1]{>{\raggedleft\let\newline\\\arraybackslash\hspace{0pt}}m{#1}}

\colingfinalcopy 

\setlist{nosep}

\newcommand\methodname{JLSD}

\title{A Joint Learning Approach based on Self-Distillation for Keyphrase Extraction from Scientific Documents}

\author{Tuan Manh Lai \textsuperscript{1}, Trung Bui \textsuperscript{2}, Doo Soon Kim \textsuperscript{2}, Quan Hung Tran \textsuperscript{2} \\
\textsuperscript{1} University of Illinois at Urbana-Champaign\\
\textsuperscript{2} Adobe Research, San Jose, CA
}

\date{}

\begin{document}
\maketitle
\begin{abstract}
\input{abstract}
\end{abstract}

%
%
\blfootnote{
    %
    %

    %
    
    
    
    \hspace{-0.65cm}  
    This work is licensed under a Creative Commons 
    Attribution 4.0 International License.
    License details:
    \url{http://creativecommons.org/licenses/by/4.0/}.
}

\section{Introduction}
\input{introduction}
\section{Preliminaries}\label{sssec:preliminaries}
\input{Preliminaries/problem_formulation}
\input{Preliminaries/baseline_model}
\section{Joint Learning based on Self-Distillation (\methodname)} \label{sssec:method}
\input{method}
\section{Experiments and Results}\label{sssec:experiments_and_results}
\input{Experiments and Results/data_and_experiment_setup}
\input{Tables/overall_results}
\input{Tables/results_unsupervised_methods}
\input{Experiments and Results/compare_previous_approaches}
\input{Experiments and Results/compare_other_joint_training_approaches}

\section{Conclusion}\label{sssec:conclusion}
\input{conclusion.tex}

\bibliographystyle{coling}
\bibliography{coling2020}

\end{document}

%% file: abstract.tex
Keyphrase extraction is the task of extracting a small set of phrases that best describe a document. Most existing benchmark datasets for the task typically have limited numbers of annotated documents, making it challenging to train increasingly complex neural networks. In contrast, digital libraries store millions of scientific articles online, covering a wide range of topics. While a significant portion of these articles contain keyphrases provided by their authors, most other articles lack such kind of annotations. Therefore, to effectively utilize these large amounts of unlabeled articles, we propose a simple and efficient joint learning approach based on the idea of self-distillation. Experimental results show that our approach consistently improves the performance of baseline models for keyphrase extraction. Furthermore, our best models outperform previous methods for the task, achieving new state-of-the-art results on two public benchmarks: Inspec and SemEval-2017.

%% file: introduction.tex
Keyphrase extraction is the task of automatically extracting a set of representative phrases from a document that concisely describe its content. As keyphrases provide a brief yet precise description of a document, they can be utilized for various downstream applications \cite{d2005keyphrase,Litvak2008GraphBasedKE,Kim2010SemEval2010T5,Boudin2013KeyphraseEF}. Over the past years, researchers have proposed many methods for the task, which can be divided into two major categories: supervised \cite{sterckx-etal-2016-supervised,zhang2017mike,alzaidy2019bi} and unsupervised techniques \cite{florescu-caragea-2017-positionrank,Boudin2018UnsupervisedKE,mahata-etal-2018-key2vec}. In the presence of sufficient domain-specific labeled data, supervised keyphrase extraction methods are often reported to outperform unsupervised methods \cite{kim2013automatic,caragea-etal-2014-citation,sahrawat2020keyphrase}.

Recently, many deep learning based methods have achieved promising performance in a wide range of NLP tasks \cite{laietal2018review,tranetal2018context,peters-etal-2018-deep,devlin-etal-2019-bert,lewis2019bart,lai2020simple}. However, most existing benchmark datasets for keyphrase extraction typically have limited numbers of annotated documents, making it challenging to train an effective deep learning model for the task. 
In contrast, digital libraries store millions of scientific articles online, covering a wide range of topics. While a significant portion of these articles have author-provided keyphrases, most other articles lack such kind of annotations. For example, major NLP conferences (i.e., ACL, EMNLP, COLING) normally do not require authors to provide keywords of their publications. In this paper, to effectively utilize these large amounts of unlabeled articles available online, we propose a novel joint learning approach based on the idea of self-distillation \cite{furlanello2018born}. To evaluate the effectiveness of our method, we use the Inspec and SemEval-2017 datasets. Experimental results show that our approach consistently improves the performance of baseline models. Furthermore, our best models achieve new state-of-the-art results on the two public benchmarks.

In summary, we have made the following contributions: (1) We propose a novel joint learning framework based on self-distillation for improving keyphrase extraction systems (2) Experiments on two public datasets, Inspec and SemEval-2017, demonstrate the effectiveness of our proposed methods. In the following parts, we first describe some preliminaries relating to the formulation of the keyphrase extraction problem and the architecture of our baseline models (Section \ref{sssec:preliminaries}). We then go into details at our self-distillation approach in Section \ref{sssec:method}. After that, we describe the conducted experiments and their results in Section \ref{sssec:experiments_and_results}. Finally, we conclude this work in Section \ref{sssec:conclusion}.

%% file: Preliminaries/problem_formulation.tex
\paragraph{Problem Formulation} Similar to recent works \cite{AAAI1714628,sahrawat2020keyphrase}, we formulate keyphrase extraction as a sequence labeling task. Let $D = (t_1, t_2, ..., t_n)$ be a document consisting of $n$ tokens, where $t_i$ represents the $i^{th}$ token of the document. The task is to predict a sequence of labels $\textbf{y} = (y_1, y_2, ..., y_n)$, where $y_i \in \{\texttt{I}, \texttt{B}, \texttt{O}\}$ is the label corresponding to token $t_i$. Label \texttt{B} denotes the beginning of a keyphrase, \texttt{I} denotes the continuation of a keyphrase, and \texttt{O} corresponds to tokens that are not part of any keyphrase. An advantage of this formulation is that it completely avoids the candidate generation step required in previous ranking-based approaches \cite{el2009kp,BennaniSmires2018SimpleUK}. Instead of having to generate a list of candidate phrases and then ranking them, we directly predict the target outputs in one go. This formulation also provides a unified approach to keyphrase extraction, as it has the same format as other sequence labeling tasks.

%% file: Preliminaries/baseline_model.tex
\begin{figure}[!t]
\centering
\includegraphics[width=0.7\textwidth]{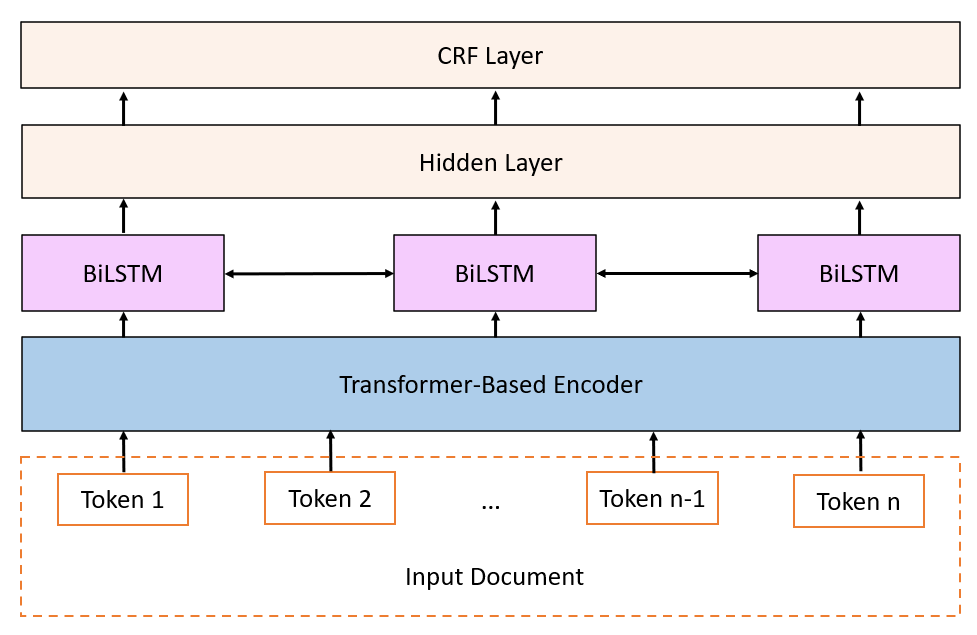}
\caption{A high-level overview of our baseline model.}
\label{fig:baseline_overview}
\end{figure}

\paragraph{Baseline Models}
In this work, we employ the BiLSTM-CRF architecture as the baseline architecture \cite{huang2015bidirectional,alzaidy2019bi,sahrawat2020keyphrase,zhu2020deep}. Figure \ref{fig:baseline_overview} shows a high-level overview of our baseline model. Given a sequence of input tokens, the model first forms a contextualized representation for each token using a Transformer-based encoder. The model then further uses a bidirectional LSTM \cite{Hochreiter1997LongSM} on top of the Transformer-based representations. After that, a dense layer is used to map the output of the bidirectional LSTM to the label space. Finally, a linear-chain CRF is applied to decode the labels.

%% file: method.tex
\begin{figure}[!t]
\centering
\includegraphics[width=0.95\textwidth]{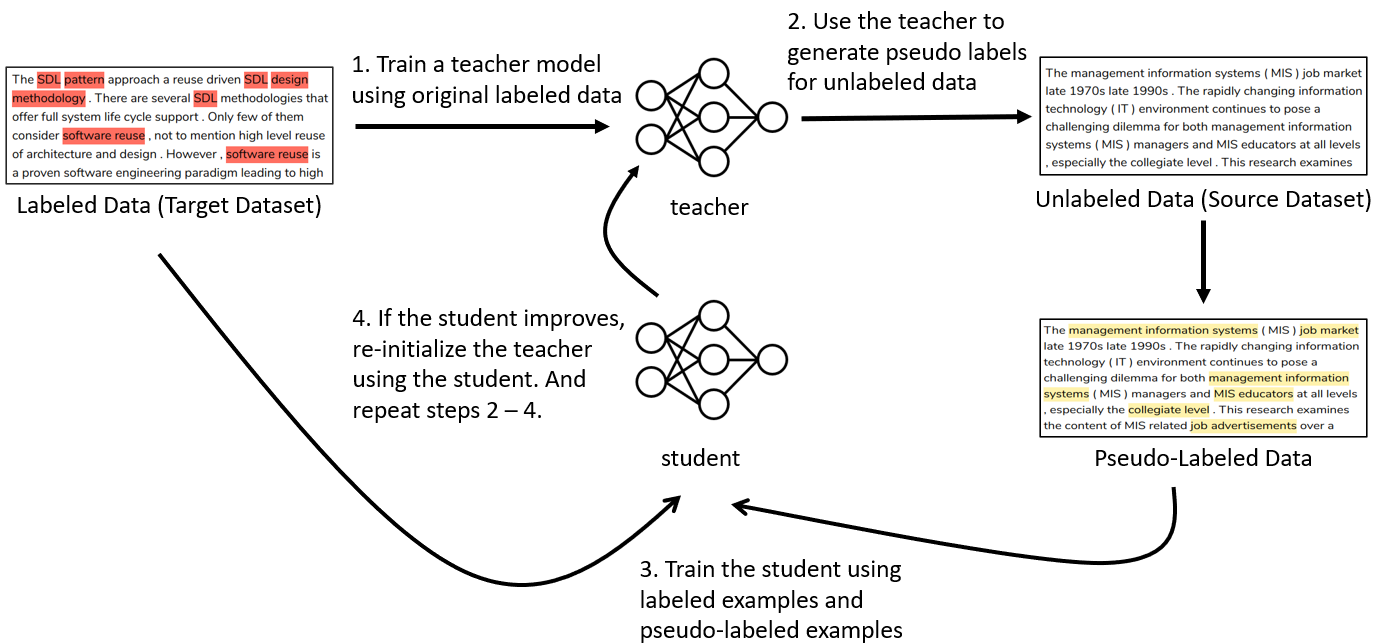}
\caption{A high-level overview of our proposed method.}
\label{fig:method_overview}
\end{figure}

Figure \ref{fig:method_overview} shows a high-level overview of our proposed self-distillation approach. We refer to the labeled dataset as the target dataset and the unlabeled dataset as the source dataset. We first train a teacher model using existing labeled examples. After that, we start training a student model parameterized identically to the teacher model. During each training iteration, we sample a batch of both original labeled examples and unlabeled examples with pseudo-labels generated by the teacher. At any point during the training process, if the student model's performance improves  (i.e., achieves better results on the validation set of the target dataset), we re-initialize the teacher model with the current parameters of the student model, and then continue training the student using the same procedure. The intuition is to produce a virtuous cycle in which both the teacher and the student will become better together. A better teacher will generate more accurate pseudo-labels, which in turn helps train a better student. And then the improved student will be used to re-initialize the teacher. Algorithm \ref{alg:main_alg} formally describes our proposed approach. Note that $T$ denotes the number of training iterations, and $r$ is a hyperparameter that determines how many unlabeled documents to be sampled in each iteration. During an iteration of \methodname, if the teacher happens to generate bad pseudo-labels for an example, the effect will not be detrimental. By design, the performance of the teacher is monotonically non-decreasing with time, therefore, when the teacher comes across the same example again, it is like to generate better pseudo-labels. To the best of our knowledge, the only other work exploring self-learning for keyphrase extraction is that of \newcite{zhu2020deep}. However, during each training epoch, their method needs to label all unlabeled examples and then selects the ones with high confidence scores. This would incur a lot of overhead for every single training epoch as the number of unlabeled examples is typically very large. Our approach does not suffer from these issues. Our teacher model generates pseudo labels on-the-fly during each training iteration. Another highly related work is the paper by \newcite{Ye2018SemiSupervisedLF}. However, the work focuses on the task of keyphrase generation instead of keyphrase extraction.

\begin{algorithm}[!h]
\SetAlgoLined
\caption{Joint Learning based on Self-Distillation (JLSD)}
\textbf{Input:} Labeled documents \big\{$(D_1, \textbf{y}_1), ..., (D_n, \textbf{y}_n)$\big\} and unlabeled documents \big\{$\widetilde{D_1},\widetilde{D_2}, ..., \widetilde{D_m}$\big\}\\
Train a teacher model using labeled documents.\\
Initialize a student model with same architecture and parameters as the teacher.\\
\For{$i = 1\;\dots\;T$}{
    Sample a batch of labeled documents $L$ uniformly at random.\\
    Sample a batch of unlabeled documents $\widetilde{U} =$ \big\{$\widetilde{D_{i1}}, ..., \widetilde{D_{ik}}$\big\} uniformly at random ($k = r|L|$).\\
    Use the teacher to generate (hard) pseudo labels for $\widetilde{U}$ to get $U =$ \big\{$(\widetilde{D_{i1}}, \textbf{y}_{i1}), ..., (\widetilde{D_{ik}}, \textbf{y}_{ik})$\big\}.\\
    Use gradient descent to update the parameters of the student using examples in $L \cup U$.\\
    \textbf{If} \big(\textit{Student performance has improved}\big) \textbf{then}
        Re-initialize the teacher using the student.
    
}
\label{alg:main_alg}
\end{algorithm}
\vspace*{-0.7cm}

%% file: Experiments and Results/data_and_experiment_setup.tex
\paragraph{Data and Experiments Setup} In this work, we experimented with two target datasets: \textbf{Inspec} and \textbf{SemEval-2017}. The \textbf{Inspec} dataset \cite{Hulth2003ImprovedAK} has 1000/500/500 abstracts of scientific articles for the train/dev/test split.  The \textbf{SemEval-2017} dataset  \cite{augenstein-etal-2017-semeval} has 350/50/100 scientific articles for the train/dev/test split. In our experiments, we use the \textbf{KP20k} dataset \cite{meng-etal-2017-deep} as the source dataset, because it contains more than 500,000 articles collected from various online digital libraries. Even though each article in the dataset has author-provided keyphrases, our proposed method does not require such supervised signals. We implemented two baseline models (Section \ref{sssec:preliminaries}) with different pre-trained contextual embeddings: BERT (base-cased)\footnote{\url{https://huggingface.co/bert-base-cased}} and SciBERT (scivocab-cased)\footnote{\url{https://huggingface.co/allenai/scibert_scivocab_cased}} \cite{Wolf2019HuggingFacesTS}.  From this point, we will refer to baseline models trained only on labeled data as [BERT] and [SciBERT]. We refer to models trained using our joint learning approach as [BERT + \methodname] and [SciBERT + \methodname]. For each variant, two different learning rates are used, one for the lower pretrained Transformer encoder and one for the upper layers. The optimal hyperparameter values are variant-specific, and we experimented with the following range of possible values: $\{4, 8, 16\}$ for batch size, $\{$2e-5, 3e-5, 4e-5, 5e-5$\}$ for lower learning rate, $\{$1e-4, 2e-4, 5e-4, 1e-3, 5e-3$\}$ for upper learning rate, and $\{25, 50, 75, 100, 125\}$ for number of training epochs. For \methodname, we experimented with various values for $r$ among $\{0.25, 0.5, 1, 1.5, 2, 4\}$. We did hyper-parameter tuning using the provided dev sets.

%% file: Tables/overall_results.tex
\begin{table}[]
\centering
\resizebox{0.75\textwidth}{!}{%
\begin{tabular}{|c|l|c|c|}
\hline
\multicolumn{1}{|l|}{\multirow{2}{*}{}} & \multirow{2}{*}{Approach} & \multicolumn{2}{c|}{Datasets} \\ \cline{3-4} 
\multicolumn{1}{|l|}{} &  & Inspec (F1) & SemEval-2017 (F1) \\ \hline
\multirow{4}{*}{Previous Models} & ELMo \cite{sahrawat2020keyphrase} & 0.568 & 0.504 \\ \cline{2-4} 
 & RoBERTA \cite{sahrawat2020keyphrase} & 0.595 & 0.508 \\ \cline{2-4} 
 & SciBERT \cite{sahrawat2020keyphrase} & 0.593 & 0.521 \\ \cline{2-4} 
 & BERT \cite{sahrawat2020keyphrase} & 0.591 & 0.522 \\ \hline\hline
\multirow{2}{*}{SciBERT-based Models} & SciBERT (Our implementation) & 0.598 $\pm$ 0.006 & 0.544 $\pm$ 0.008 \\ \cline{2-4} 
 & SciBERT + \methodname & \textbf{0.605 $\pm$ 0.001} & 0.546 $\pm$ 0.002 \\ \hline\hline
\multirow{4}{*}{BERT-based Models} & BERT (Our implementation) & 0.596 $\pm$ 0.003 & 0.537 $\pm$ 0.003 \\ \cline{2-4} 
 & BERT + Simple Joint Training & 0.582 $\pm$ 0.004 & 0.545 $\pm$ 0.008 \\ \cline{2-4} 
 & BERT + Simple Pretraining & 0.588 $\pm$ 0.002 & 0.548 $\pm$ 0.004 \\ \cline{2-4} 
 & BERT + \methodname & 0.601 $\pm$ 0.004 & \textbf{0.554 $\pm$ 0.004} \\ \hline
\end{tabular}%
}
\caption{Overall results on the target datasets (means and standard deviations of our results are shown).}
\label{tab:overall-results}
\end{table}

%% file: Tables/results_unsupervised_methods.tex
\begin{table}[]
\centering
\resizebox{0.75\textwidth}{!}{%
\begin{tabular}{|c|c|c|c|c|c|c|c|c|c|c|c|}
\hline
\multirow{2}{*}{Metrics} & \multicolumn{4}{c|}{Our Models} & \multicolumn{7}{c|}{Unsupervised Ranking-based Models} \\ \cline{2-12} 
 & \begin{tabular}[c]{@{}c@{}}BERT +\\ JLSD\end{tabular} & \multicolumn{1}{l|}{BERT} & \begin{tabular}[c]{@{}c@{}}SciBERT + \\ JLSD\end{tabular} & \multicolumn{1}{l|}{SciBERT} & SIF & Embed & RVA & Position & Topic & Single & \multicolumn{1}{l|}{YAKE} \\ \hline
F1@5 & 42.06 & 41.97 & \textbf{43.32} & 42.39 & 29.11 & 27.20 & 21.91 & 25.19 & 22.76 & 24.69 & 15.73 \\ \hline
F1@10 & 57.52 & 57.15 & \textbf{58.05} & 57.25 & 38.80 & 34.98 & 30.13 & 31.53 & 27.30 & 33.19 & 19.07 \\ \hline
F1@15 & 66.05 & 66.01 & \textbf{66.88} & 66.13 & 39.59 & 36.45 & 33.22 & 33.30 & 28.42 & 35.42 & 20.17 \\ \hline
\end{tabular}%
}
\caption{Comparison of our models and unsupervised models (F1@5, F1@10, and F1@15 test scores on the Inspec dataset are reported). To save space, standard deviations of our results are not shown.}
\label{tab:results_unsupervised_methods}
\end{table}

%% file: Experiments and Results/compare_previous_approaches.tex
\paragraph{Comparison with Previous Supervised Methods} We first compare the performance of our models with the supervised models proposed by \newcite{sahrawat2020keyphrase}, as they have recently achieved state-of-the-art results on the two target datasets. Our baseline models are similar to the models proposed by \newcite{sahrawat2020keyphrase}, each consists of a contextualized embedding layer followed by a BiLSTM-CRF structure. Table \ref{tab:overall-results} presents the overall results on the Inspec and SemEval-2017 datasets. Results of our models are averaged over three random seeds. When calculating the F1 scores, we consider only extractive keyphrases that are present in the documents. We see that our own implementation of the [BERT] and [SciBERT] variants achieve better results than reported by \newcite{sahrawat2020keyphrase}. Also, by applying our proposed joint learning method (\methodname), we can consistently improve the performance of the baseline variants. For example, the [SciBERT + \methodname] variant achieves new state-of-the-art performance on the Inspec dataset, while the [BERT + \methodname] also outperforms previous methods on the SemEval-2017 dataset. These results demonstrate the effectiveness of our joint learning approach. 

\paragraph{Comparison with Previous Unsupervised Methods} We also compare our models with previous state-of-the-art unsupervised approaches, including SIFRank \cite{Sun2020SIFRankAN}, EmbedRank \cite{BennaniSmires2018SimpleUK}, RVA \cite{papagiannopoulou2018local},  PositionRank \cite{Florescu2017APP}, TopicRank \cite{Bougouin2013TopicRankGT}, SingleRank \cite{Wan2008SingleDK}, and YAKE \cite{campos2018yake}. Table \ref{tab:results_unsupervised_methods} presents the comparison on the Inspec dataset. In this case, since the unsupervised methods are ranking-based methods, the performances are evaluated in terms of F1-measure when a fixed number of top keyphrases are extracted (i.e., F1@5, F1@10, and F1@15 measures are used). We see that our baseline models [BERT] and [SciBERT] already outperform previous unsupervised methods by a large margin. This agrees with previous studies, suggesting that in the presence of sufficient labeled data, supervised methods typically perform better than unsupervised methods \cite{kim2013automatic,caragea-etal-2014-citation}. Also, by applying \methodname, we further increase the gap with previous unsupervised methods. Note that, by default, our baseline model is a sequence labeling model (Section \ref{sssec:preliminaries}). Therefore, to compare our models with previous unsupervised ranking methods, we need to convert our models into ranking models by deriving a way to compute confidence scores for predicted keyphrases. In order to do so, we first calculate marginal probabilities after the CRF layer and simply make an independence assumption to compute the probability for a predicted keyphrase.

%% file: Experiments and Results/compare_other_joint_training_approaches.tex
\paragraph{Comparison with Other Transfer Learning Techniques} Finally, we compare our joint learning approach to other popular transfer learning techniques, including \textit{simple pretraining} and \textit{simple joint training}. Despite their simplicity, these strategies have been shown to be effective for a wide range of tasks \cite{minetal2017question,lai2018supervised,yoon2019compare,laietal2019gated,langenfeld2019protein}. In \textit{simple pretraining}, we first train a baseline model on the scientific articles in the source dataset. After that, we simply finetune the same model on a target dataset (i.e., Inspec or SemEval-2017). In \textit{simple joint training}, we train a baseline model using examples from both the source dataset and the target dataset at the same time. Before each new training epoch, we sample a new set of examples from the source dataset and add them to the pool of examples of the target dataset. Different from our proposed approach, these two transfer learning techniques are only applicable to situations where the source dataset is labeled. In this case, since each article in the source dataset (i.e., KP20k) contains author-provided keyphrases, we use these keyphrases as the supervised signals. From the Table \ref{tab:overall-results}, we see that our variant [BERT + \methodname] outperforms both variants [BERT + Simple Joint Training] and [BERT + Simple Pretraining], even though [BERT + \methodname] does not require the source dataset to have any supervised signal. Furthermore, we see that the two simple transfer learning approaches even adversely decrease the performance of the BERT baseline model on the Inspec dataset.

%% file: conclusion.tex
In this work, we propose a novel joint learning approach based on self-distillation. Experimental results show that our approach consistently improves the performance of baseline models. Our best models even achieve new state-of-the-art results on two public benchmarks (Inspec and SemEval-2017). In future work, we plan to explore how to extend our method to other tasks \cite{brixey2018system} and other languages. We will also investigate how our method can be used in few-shot settings.